\documentclass{article} 
\usepackage{iclr2021_conference,times}

\usepackage{amsmath,amsfonts,bm}

\def\eqref#1{equation~\ref{#1}}

\def\1{\bm{1}}

\def\ry{{\textnormal{y}}}

\def\vx{{\bm{x}}}
\def\vy{{\bm{y}}}

\def\mI{{\bm{I}}}

\def\mW{{\bm{W}}}

\DeclareMathAlphabet{\mathsfit}{\encodingdefault}{\sfdefault}{m}{sl}
\SetMathAlphabet{\mathsfit}{bold}{\encodingdefault}{\sfdefault}{bx}{n}

\def\sM{{\mathbb{M}}}

\def\emW{{W}}

\usepackage{hyperref}
\usepackage{url}

\usepackage{graphicx}

\title{Vision-based route following by an embodied insect-inspired sparse neural network}

\author{Lu Yihe, Rana Alkhoury Maroun \& Barbara Webb \\
School of Informatics\\
University of Edinburgh\\
Edinburgh, Scotland, UK \\
\texttt{\{yihe.lu,r.alkhoury-maroun,b.webb\}@ed.ac.uk} 
}

\iclrfinalcopy 
\begin{document}

\maketitle

\begin{abstract}
We compared the efficiency of the FlyHash model, an insect-inspired sparse neural network \citep{dasgupta2017neural}, to similar but non-sparse models in an embodied navigation task. 
This requires a model to control steering by comparing current visual inputs to memories stored along a training route.
We concluded the FlyHash model is more efficient than others, especially in terms of data encoding.
\end{abstract}

\section{Introduction}
\begin{figure}[h]
\begin{center}
\centerline{\includegraphics[width=0.6\textwidth]{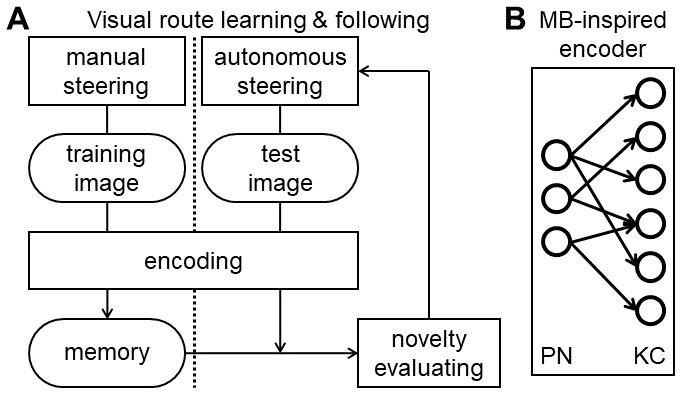}}
\end{center}
\caption{
{\bf A}: A diagram of an embodied model for vision-based route following.
Visual memory is stored only in training sessions, and retrieved for novelty evaluation only in tests.
{\bf B}: An insect mushroom body (MB)-inspired neural network.
The projection neurons (PNs) receive sensory inputs, and pass them onto the Kenyon cells (KC) for data encoding.
In the real MB structure, and typically in the FlyHash model, 
the number of KCs is larger than the number of PNs, i.e., $\displaystyle N_\text{KC} > N_\text{PN}$,
and their connectivity is sparse, 
as is the proportion of KCs activated for encoding. 
Here, we also explore dense connectivity and non-sparse coding. 
}
\label{fig:mb}
\end{figure}

The ability of animals to follow a familiar route is one of their critical navigation skills, which enables them to forage, to escape, to home, or to migrate, without exploring unknown, potentially dangerous, territories.
Despite their tiny brains, desert ants are capable of vision-based route following.
Based on a brain structure, called mushroom body (MB), which is common across many insect species,
\citet{ardin2016using} developed a computational model, which can learn a training route in a one-shot manner, and subsequently follow the route reliably.
This MB-based model is fundamentally a sparse, shallow neural network performing unsupervised learning of retinotopic images.
After training, it succeeds in route following by selecting the direction associated with the most familiar visual input (Figure \ref{fig:mb}A).

Inspired by the same insect MB structure, \citet{dasgupta2017neural} proposes the FlyHash model, which performs locality-sensitive hashing (LSH).
In contrast to classical hashing which aims to minimise encoding duplication amongst inputs (even if two inputs are similar),
LSH algorithms assign similar hashes to similar inputs
to support efficient similarity search in large databases by approximating nearest neighbour search \citep{andoni2008near}.
Since typical LSH hashes have smaller dimensionality than their inputs, LSH can be considered as a dimensionality reduction technique.
Unlike other techniques that exploit data extensively, e.g., to train an autoencoder, to compute orthonormal basis in principle component analysis, 
before they can be used for dimensionality reduction,
LSH algorithms take advantage of randomly initialised functions to encode inputs as hashes.
Typically, such encoding functions stay unchanged and data-independent after initialisation.

The FlyHash model effectively formalises the visual encoding process previously used in the MB-based route following model,
as the same neural network is considered (Figure \ref{fig:mb}B).
A sensory signal received by the input layer, composed of projection neurons (PNs), is encoded as a hash, represented by binarised activation of Kenyon cells (KCs).
Two features of this MB-inspired neural network
make the FlyHash model distinct from conventional LSH algorithms.
Firstly, an input is encoded as a higher-dimensional sparse hash,
Secondly, the PN-KC connectivity, defining the hash function, is sparse and binary.

Since it is beneficial for an intelligent agent, either biological or artificial, to encode and store memory efficiently and effectively for visual recognition of a familiar route, it might seem the smaller hash produced by 
a conventional LSH technique would be more appropriate for such tasks.
Nevertheless, the FlyHash model suggests an alternative, seemingly counter-intuitive solution, which appears to be what is used by the real insect MB.
Therefore, we deployed the FlyHash model on a virtual robot in a realistic simulation environment, along with two alternative encoding algorithms, in order to compare their model efficiency in a task-relevant manner.

\section{Methods}

\subsection{Embodied models}
In this manuscript the embodied models are composed of 3 components, a data encoder, a memory storage, and a steering controller (Figure \ref{fig:mb}A).
As we compare only the models for data encoding, while keeping the other 2 components identical, we use the same name for an embodied robot model and the model for date encoding deployed on the robot.

\subsection{Models for data encoding}
\label{sec:main-model}
We consider 3 single-layer neural network models (same or similar to Figure \ref{fig:mb}B) for data encoding,
and their encoding functions can be formulated as,
\begin{equation}
\displaystyle
    \vy = f_{\mW}(\vx) = f(\mW \cdot \vx),
\end{equation}
where $\displaystyle \vx$ denotes an input vector of length $\displaystyle N_\text{PN}$, and $\displaystyle \vy$ the encoded output of dimension $\displaystyle N_\text{KC}$.
Note that the activation function $\displaystyle f$, the connection matrix $\displaystyle \mW$ and the encoding matrix $\displaystyle f_{\mW}(\vx)$ are model specific, but not plastic.

{\bf The FlyHash model} is characterised by its sparse and binary $\displaystyle \mW$ forming an expansion structure, i.e., $\displaystyle N_\text{PN} < \displaystyle N_\text{KC}$ \citep{dasgupta2017neural}.
$\displaystyle \mW$ is initialised by randomly generating its weights independently and identically from a Bernoulli distribution, $\displaystyle \emW_{ij} \sim \mathcal{B}(\theta)$,
where $\displaystyle i \in \{1, 2, \dots, N_\text{PN} \}, ~j \in \{1, 2, \dots, N_\text{KC} \}$, and $\displaystyle \theta \in (0, 1)$ controls the sparsity of $\displaystyle \mW$.
$\displaystyle f$ performs $k$-winners-take-all ($k$-WTA),
which binarises $\displaystyle \mW \cdot \vx$,
and produces $\displaystyle \vy$ of sparsity, $\displaystyle \kappa = k / N_\text{KC} \in (0, 1)$.

{\bf The conventional LSH model} typically uses dense and non-binary $\displaystyle \mW$ \citep{indyk1998approximate}.
It is initialised by randomly generating its weights independently and identically from a standard Gaussian distribution, $\displaystyle \emW_{ij} \sim \mathcal{N}(0;1)$. $\displaystyle f$ is essentially the Heaviside step function.
Consequently, conventional LSH hashes are binary but not sparse ($\displaystyle \kappa=0.5$ on average).

{\bf The perfect memory model} does not encode inputs.
Instead, veridical inputs are stored in training sessions,
and evaluated in tests,
i.e., $\displaystyle \vy = \vx$.
For consistency, we can still define an identity operator, $\displaystyle f_{\mW} = \mI$, to be the model's encoding function.
As in \citet{ardin2016using}, the perfect memory model serves as a benchmark model in the route following task.

\subsection{Memory storage and novelty evaluation}
\label{sec:storage-and-novelty}
While the perfect memory model simply stores visual inputs without any encoding,
both the FlyHash model and the conventional LSH model store hashes only.
Consequently, to evaluate visual novelty $\displaystyle d$, the dissimilarity metric $\displaystyle D$ is either the Hamming distance for binary hashes, or the Euclidean distance for images.
Since there are multiple memory items stored from a training session, the visual novelty of an input is computed as
\begin{equation}
\label{eq:novelty}
\displaystyle
    d(\vx) = \min_{\vy \in \sM } D \left( f_{\mW}(\vx), \vy \right),
\end{equation}
where $\displaystyle \sM$ is the set of all stored memory items.
According to the theory of LSH, two sufficiently similar images should share the same hash, given the same $\displaystyle f_{\mW}$.

The same number of memory items is considered for the 3 models,
so that differences in their behaviour and performance are determined mainly by the quality of memory encoding,
and their model size only by the size of individual memory items.

\subsection{Insect-inspired model for autonomous robot steering}
\label{sec:steering}
Using a well-established insect central complex model \citep{stone2017anatomically},
which accounts for insect steering (and other navigation skills) by an anatomically realistic neural network,
\citet{wystrach2020lateralised} shows that insect locomotion can be robustly stabilised to follow straight trajectories. 
The key idea is that novelty evaluation (treated as an idealised function in \citet{wystrach2020lateralised}) occurs in a pair of MB structures with their inputs lateralised to left and right respectively, thus their relative outputs inform the central complex model of the correct turning direction to bring the insect back in line with the route.

Here we use a functionally similar but structurally simpler motor controller for autonomous robot steering,
which modulates the locomotion direction based on realistic lateralised visual inputs (see Section \ref{sec:igibson} for our implementation).
Specifically, a data encoding model stores visual memory in training sessions,
and at any moment of autonomous control it computes left and right visual novelty $\displaystyle d_\text{L}$ and $\displaystyle d_\text{R}$ of left and right visual inputs by Equation (\ref{eq:novelty}).
The motor controller then determines the robot's angular speed $\displaystyle \omega$ according to the novelty difference.
Specifically,
\begin{equation}
\label{eq:LRcontrol}
\displaystyle
    \omega = \alpha \cdot \frac{d_\text{L} - d_\text{R}}{d_\text{L} + d_\text{R}},
\end{equation}
where $\displaystyle \alpha > 0$ is a predetermined constant. 
The normalisation term $\displaystyle d_\text{L} + d_\text{R}$ is used to remove the effect of the number of learned views.

\section{Experiment}
\label{sec:experiment}

\subsection{Simulation environment: iGibson}
\label{sec:igibson}
\begin{figure}[h]
\begin{center}
\centerline{\includegraphics[width=0.6\textwidth]{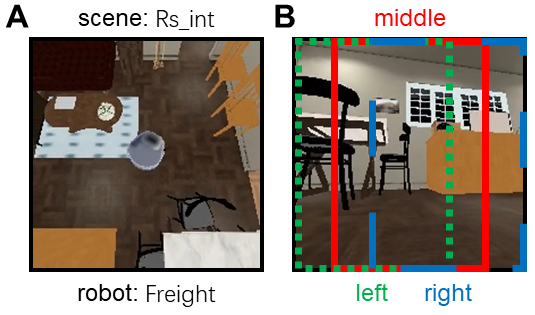}}
\end{center}
\caption{Snapshots of the {\tt Freight} robot running in the {\tt Rs\_int} scene.
{\bf A}: A bird's-eyed view. 
{\bf B}: A raw image taken by the robot camera.
Visual inputs from the middle field (bounded by the red, solid lines) were stored by the models, 
visual inputs from the left field (bound by the green, dotted lines) and the right field (bounded by the blue, dashed lines) were used for computing visual novelty. 
}
\label{fig:robot}
\end{figure}

The simulated experiments were conducted in iGibson, an environment offering realistic visual rendering and physics simulation \citep{shenigibson}.
Our models were deployed on the two wheeled {\tt Freight} robot,
which was running in the {\tt Rs\_int} scene (Figure \ref{fig:robot}A).
While there is built-in noise in the simulated physics of iGibson, its default value caused negligible effects on our simulations.

Raw image were taken by the RGB-D camera on the robot was downsampled to the size of $33\times 33$ pixels, transformed to greyscale by the standard colour converting method from the OpenCV library, and blurred by averaging pixels in every $7\times 7$-sized boxes.
The depth information was discarded.
We found empirically that such grey and small (but not too small) images were sufficient for our models to achieve good performance,
while permitting faster and cheaper simulations. Such low resolution vision is also consistent with insect compound eye optics.

The image was cropped into left, middle and right visual fields (Figure \ref{fig:robot}B).
The sizes of the 3 fields were identically $22\times 33$ pixels. The middle field was used as the input to be stored in visual memory during training. 
The left and the right fields were used as inputs in test sessions for computing visual novelty, as described in Section \ref{sec:steering}.

The robot was controlled by commands in terms of linear speed $\displaystyle v$ and angular speed $\displaystyle \omega$ of its wheel joints.
While $\displaystyle v$ and $\displaystyle \omega$ was manually specified in training sessions,
$\displaystyle \omega$ was determined by our models autonomously using Equation (\ref{eq:LRcontrol}) with $\alpha = 1$ in tests.
For simplicity, $\displaystyle v$ in the training and the test sessions were set to be constants, $\displaystyle v_\text{train}=0.5$ and $\displaystyle v_\text{test}=0.2$.\footnote{
We note that the unit of the robot's linear speed (omitted elsewhere for brevity) is $\displaystyle 2\pi \times 0.0613 \approx 0.3852$ meter per second,
as the robot's wheel is $0.0613$ meter in radius.}
Since the robot deviated more from the training route when it moved at higher speeds, 
we set $\displaystyle v_\text{test}=0.2$ to ensure that the model performance depends only on the effectiveness of the data encoding. 

\subsection{Model specification}
\label{sec:parameter}
The number of PNs was the same as the number of pixels, i.e., $\displaystyle N_\text{PN} = 726$,
and the values in an input vector $\displaystyle \vx$ were simply the greyscale values of the pixels.
We varied the number of KCs, i.e., the hash length $\displaystyle N_\text{KC}$, to compare models of different sizes (summarised in Table \ref{tab:model-size}),
because the overall model size is asymptotically dominated by $\displaystyle N_\text{KC}$, as a model stores more memory items $\displaystyle \vy$, .

\begin{table}[t]
\caption{Model comparison in terms of data storage size}
\label{tab:model-size}
\begin{center}
\begin{tabular}{cllll}
&
&\multicolumn{1}{c}{\bf FlyHash}
&\multicolumn{1}{c}{\bf conventional LSH}  
&\multicolumn{1}{c}{\bf perfect memory}
\\ \hline 
& data type 
& binary
& real ({\tt float})
& N/A \\
$\displaystyle \mW$
& dimension
& $\displaystyle N_\text{PN} \times N_\text{KC}$ (sparse) 
& $\displaystyle N_\text{PN} \times N_\text{KC}$ (all-to-all)
& N/A \\
& {\bf size} (bit)
& $\displaystyle N_\text{PN} \times N_\text{KC}$
& $\displaystyle 64 \times N_\text{PN} \times N_\text{KC}$
& $0$ \\
\hline  
& data type 
& binary
& binary
& greyscale ({\tt uint8}) \\
$\displaystyle \vy$
& dimension
& $\displaystyle N_\text{KC}$ (sparse) 
& $\displaystyle N_\text{KC}$ (dense) 
& $\displaystyle N_\text{PN}$ (dense) \\
& {\bf size} (bit)
& $\displaystyle N_\text{KC}$ 
& $\displaystyle N_\text{KC}$
& $\displaystyle 8 \times N_\text{PN}$ \\
\hline

\end{tabular}
\end{center}
\end{table}

In the FlyHash model, the PN-KC connectivity matrix $\displaystyle \mW$ was sparse with $\displaystyle \theta = 10 / N_\text{PN}$;
on average a KC was connected to $10$ PNs.
We consider 3 sparsity levels for $\displaystyle \vy$, $\displaystyle \kappa = 0.05, 0.1$ and $0.5$.
Note that the model is biologically realistic when $\displaystyle \theta = 10 / N_\text{PN}$ and $\displaystyle \kappa = 0.05$ or $0.1$ \citep{lin2014sparse}.
When a hash $\displaystyle \kappa=0.5$, $\displaystyle \vy$ was not sparse,
but as dense as that in the conventional LSH model.

In $\displaystyle \mW$ and $\displaystyle \vy$,
each binary element was assumed to occupy 1 bit,
a greyscale value 8 bits,
and a real number 64 bits.\footnote{
By default, a binary variable represented by {\tt bool} in Python occupies 8 bits, 
and iGibson processes images with 32-bit pixels,
which are larger than what we assume.
Nevertheless, our assumptions are made for theoretical analysis of model size,
and the efficient implementation of the underlying data type is beyond the scope of this work,
}

\subsection{Task: route following}
\label{sec:task}
An experimental trial was composed of a training session and a test session,
and all trials were independent from one another.
The robot was trained under manual control along a predetermined route,
which was designed arbitrarily by the authors with the only constraint that no collisions were to occur in training sessions.
25 images along the route were taken every 0.5 seconds (in simulation time), and stored by the model.

In the subsequent test session, the robot started from the same position, heading the same direction, as in the training session, but was now
under full control of the model.
The goal was to follow the training route as far as possible.
Note that a collision event could and did occur under autonomous control in a trial as the result of poor memory encoding,
and would be severely detrimental to the overall route following performance.

\section{Results}
\label{sec:result}

\subsection{Model comparison: route following performance}
\begin{figure}[h]
\begin{center}
\centerline{\includegraphics[width=\textwidth]{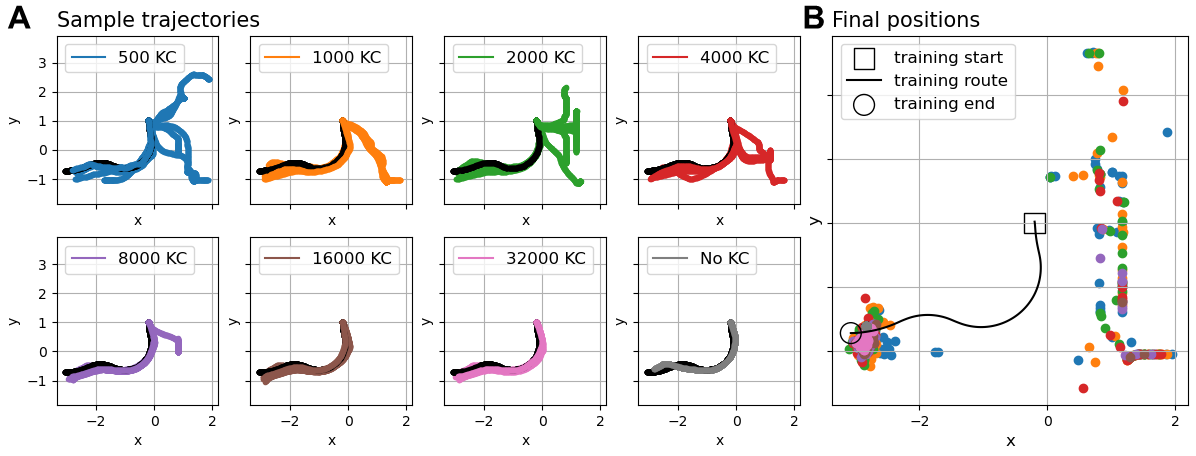}}
\end{center}
\caption{Route following behaviour of the FlyHash model with hash sparsity, $\displaystyle \kappa = 0.1$.
{\bf A}: Sample trajectories of 10 trials for different $\displaystyle N_\text{KC}$ (coloured).
In the case of no KC, the perfect memory model was used instead. 
{\bf B}: Final positions of 100 trials.
All training routes (in black) started at the same position, $\displaystyle (x, y) = (-0.20, 1.00)$, overlaid on one another due to negligible simulation noise, and ended at the same position, approximately $\displaystyle (x, y) = (-3.07, -0.72)$.
The results of the conventional LSH model were qualitatively similar, and the plot is thus omitted.
Note that it was impossible for the robot to take a shortcut between the start and the final positions due to obstacles.
A video demo of the embodied perfect memory model can be found at \url{https://youtu.be/EdaQrm1ruQQ}.
}
\label{fig:vary-kc-fh}
\end{figure}

We first checked the perfect memory model, which could reliably achieve one-shot vision-based route following.
For instance, its sample trajectories from 10 trials overlay on one another in Figure \ref{fig:vary-kc-fh}A (labelled `No KC').

It is clear in Figure \ref{fig:vary-kc-fh}A that the length of hashes $\displaystyle N_\text{KC}$ had a significant impact on the route following behaviour of the FlyHash model,
To quantify such results, we measured the final distance between the final positions of the training and the test trajectories in individual trials,
and considered a trial successful if the final distance was less than $2$ meters.
We consider this criterion a reasonable proxy measure for the overall route following performance,
because the final positions either clustered near that of the training route when the route was followed,
or distributed widely when the robot deviated too much and became lost (Figure \ref{fig:vary-kc-fh}B).

\begin{figure}[h]
\begin{center}
\centerline{\includegraphics[width=0.9\textwidth]{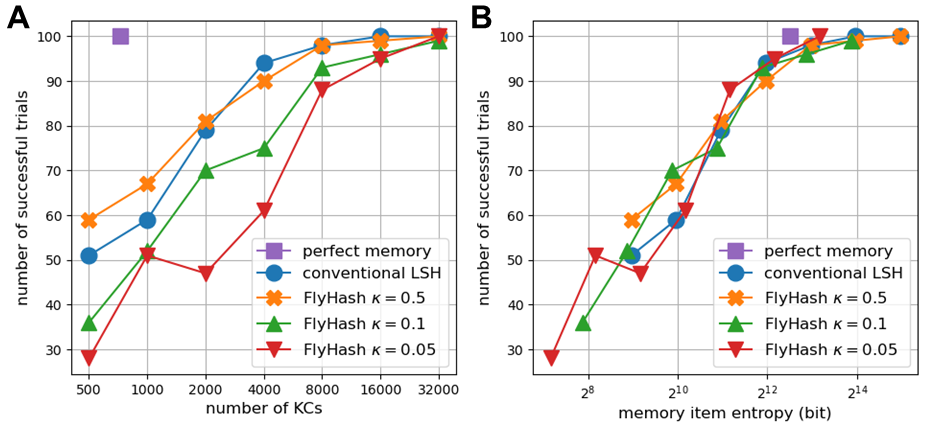}}
\end{center}
\caption{Route following performance modualated by model size.
{\bf A}: The relation between task performance and $\displaystyle N_\text{KC}$.
{\bf B}: The relation between the same task performance and the average entropy of a stored memory item.
}
\label{fig:compression}
\end{figure}

While the perfect memory model reached perfect performance effectively with $726$ 8-bit KCs (i.e., $5808$ bits in total),
the conventional LSH model and the FlyHash model with dense hashes ($\displaystyle \kappa = 0.5$) required $16000$ binary KCs,
and the FlyHash models with biologically realistic, sparse hashes ($\displaystyle \kappa = 0.05$ and $0.1$) $32000$ binary KCs, to achieve the same (Figure \ref{fig:compression}A).
If $10\%$ failure was acceptable, the models with sparse hashes would require only $8000$ KCs, and those with dense hashes $4000$ KCs,
which were more comparable to the perfect memory model in terms of the size of individual memory items.

The performance was reduced more as $\displaystyle N_\text{KC}$ decreased,
probably because a smaller $\displaystyle N_\text{KC}$ was more likely to introduce hash collisions for dissimilar inputs (violating the principle of LSH),
producing ill-computed visual novelty that disrupted the route following behaviour.
The performance deterioration seemed relatively severe for the FlyHash model with the sparsest hashes ($\displaystyle \kappa = 0.05$) for an intermediate number of KCs ($\displaystyle 2000 \leq N_\text{KC} \leq 8000$).

Interestingly, while the conventional LSH model was slightly more advantageous than the FlyHash model with $\displaystyle \kappa = 0.5$ for $\displaystyle N_\text{KC} \geq 4000$,
it was outperformed by the FlyHash model with a performance gap up to $10\%$ when $\displaystyle N_\text{KC}$ was smaller.

\subsection{Model comparison: model size and run-time cost}
The above observation of the FlyHash model is particularly intriguing to us,
because using dense hashes, rather than sparse encoding as in real insect MBs, yielded the best performance.

We note that it is possible to compress the sparse hashes, but not the dense ones, in a lossless manner by exploiting their sparsity.
According to Shannon's source coding theorem, the lower bound of the code length for compression is determined by the entropy of the source code,
\begin{equation}
\label{eq:entropy}
\displaystyle 
H(\ry) = -\sum_i p_i \log p_i,
\end{equation}
where each bit $\displaystyle \ry$ in a sparse hash in the FlyHash model followed a Bernoulli distribution, $\displaystyle \ry \sim \mathcal{B}(\kappa)$.
Therefore, a sparse hash with $\displaystyle \kappa=0.1$ or $0.05$ could be compressed in lossless manner at most to approximately $\displaystyle 0.4690 N_\text{KC}$ or $\displaystyle 0.2864N_\text{KC}$ bits, respectively,
whereas the hasehs in the FlyHash model with $\displaystyle \kappa=0.5$ or the conventional LSH model could not be losslessly compressed.
As a result, despite different $\displaystyle \kappa$, the FlyHash model could achieve a performance level similar to the conventional LSH model with comparable memory sizes (Figure \ref{fig:compression}B).
Considering additionally that its $\displaystyle \mW$ was also sparse and binary (see Table \ref{tab:model-size} for detailed comparison),
the FlyHash model could be more efficient than the conventional LSH model in terms of overall model size.

The sparse and binary $\displaystyle \mW$ further implies that fewer run-time operations were required (Table \ref{tab:operation}).
To compare the encoding costs of the FlyHash model and the conventional LSH model,
it is straightforward to refer to the time complexity of the best sorting algorithm (for $k$-WTA) and matrix multiplication (between an $\displaystyle m \times n$ matrix and an $\displaystyle n \times p$ matrix),
i.e., $\displaystyle O(n \log n)$ versus $\displaystyle O(mnp)$,
where $\displaystyle m = N_\text{PN}, n  = N_\text{KC}$ and $\displaystyle p=1$.
Since $\displaystyle \log N_\text{KC} \ll N_\text{PN}$ in all our models,
the FlyHash model is guaranteed to be more efficient than the conventional LSH model in terms of run-time cost.

\begin{table}[t]
\caption{Model comparison in terms of run-time operations}
\label{tab:operation}

\begin{center}
\begin{tabular}{lllll}
\multicolumn{1}{c}{\bf Phase}  
&\multicolumn{1}{c}{\bf Operation}  
&\multicolumn{1}{c}{\bf FlyHash}
&\multicolumn{1}{c}{\bf conventional LSH}  
&\multicolumn{1}{c}{\bf perfect memory}
\\ \hline 
& multiplication
& $0$ 
& $\displaystyle N_\text{PN} \times N_\text{KC}$
& $0$ \\
encoding
& addition
& $\displaystyle 9 \times N_\text{KC}$ 
& $\displaystyle (N_\text{PN} - 1) \times N_\text{KC}$ 
& $0$ \\
& $k$-WTA
& $1$ 
& $0$
& $0$ \\
\hline  
& XOR
& $\displaystyle N_\text{KC}$ 
& $\displaystyle N_\text{KC}$ 
& $0$ \\
evaluating
& multiplication (square)
& $0$ 
& $0$
& $\displaystyle N_\text{PN} + 1$ \\
& addition (subtraction)
& $\displaystyle \leq 2\kappa \times N_\text{KC}$ 
& $\displaystyle \leq N_\text{KC}$ 
& $2 \times N_\text{PN} - 1$ \\
\hline

\end{tabular}
\end{center}
\end{table}

Although processing input images directly allowed the perfect memory model to avoid any encoding costs, the computation of visual novelty in terms of the Euclidean distance between an images was less efficient than finding the Hamming distance between hashes in the other 2 models.

\section{Discussion}
In this work in progress, we investigated the data encoding and storage efficiency of the FlyHash model,
which uses an insect-inspired sparse neural network to encode data as sparse hashes,
by deploying it on a virtual robot in a vision-based route following task.
We compared it to the conventional LSH model,
which shares many structural and functional similarities with the FlyHash model, except using a dense neural network and dense hashes.
The route following performance of the models was positively correlated with their model sizes,
and they could achieve perfect performance given sufficiently large memory.

While the conventional LSH model outperformed the FlyHash model given the same model size, 
the FlyHash model with biologically realistic hashes could be largely but losslessly compressible due to its sparsity, 
which was impossible for the conventional LSH model.
After compression, the 2 models could indeed be comparable in model size at a similar route following performance level.
In addition, the FlyHash model with biologically unrealistic dense hashes was more efficient in terms of run-time cost than the conventional LSH model, due to its sparse and binary PN-KC connectivity.

In summary. we considered the FlyHash model generally more efficient, mainly because it required a lower run-time costs.

\subsection{Limitations and future work}
The main limitation of the current work is that our model compression analysis was based on theoretical low bound;
an explicit, efficient method to achieve the low bounds might not be easily accessible.
For instance, the compressed sparse row format would be a data structure naturally compatible with our implementation for both the sparse connectivity matrix and the sparse hashes.
The format will require storing $\displaystyle \kappa \times N_\text{KC}$ indices for every $\displaystyle \vy$, occupying $\displaystyle 0.8 N_\text{KC}$ bits if {\tt uint16} is used for the $\displaystyle \vy$ indices in the FlyHash model with $\displaystyle N_\text{KC} = 32000$ and $\displaystyle \kappa = 0.05$, much more than the theoretical lower bound ($\displaystyle 0.2864N_\text{KC}$ bits).

An alternative solution would be to store implicitly visual information in plastic neural weights in learning, instead of storing memory items explicitly and computing visual novelty with respect to all the items.
With an additional output layer of even one neuron connecting to all the KCs, it should be sufficient for our models to accomplish visual route following as in \citet{ardin2016using},
because both our approaches rely on visual novelty detection, despite the different steering models.
We are planning to add this layer,
and we expect the modified FlyHash model to remain the most efficient for 2 reasons.
Firstly, its encoding matrix is sparse and binary (as discussed in this manuscript).
Secondly, the modified model would effectively be able to store more images (implicitly),
because the memory capacity $\displaystyle m$ would be approximately
\begin{equation}
\displaystyle    
    m = \frac{\log\left(1-p_\text{error}^{1/N_\text{KC}}\right) - \log \kappa}{\log(1-\kappa)},
\end{equation}
where $p_\text{error}$ denotes the probability for the model to confuse a novel visual input mistakenly as a familiar one \citep{ardin2016using}.
Thus, with $p_\text{error} = 1\%$, the modified FlyHash model with $\displaystyle \kappa=0.05$ would merely require $330$ KCs to distinguish novel inputs from 25 training images.
Most interestingly, $\displaystyle m$ increases as $\displaystyle \kappa$ becomes smaller,
which can be explained by the formal proof that the expanded, sparse, and binary hashes can be more easily learned by an output neuron \citep{dasgupta2020expressivity}.
The modified model also reflects better the real structure of the insect MB. 
It is also sensible in terms of metabolic efficiency and functional effectiveness,
because if the KC activation were dense, the downstream synapses would have to adjust their weights rapidly and extensively.

In addition, we are interested in comparing the (modified) FlyHash model to other data encoding techniques on a virtual robot,
especially in simulations with larger physical, visual, or neural noise.
It will be even more interesting to compare them on an real embodied robot.
Such endeavours will deepen our understanding of the efficiency of biological intelligence, 
and will hopefully help us build faster and less energy demanding artificial intelligence.

\bibliography{iclr2021_conference}

\begin{thebibliography}{9}
\providecommand{\natexlab}[1]{#1}
\providecommand{\url}[1]{\texttt{#1}}
\expandafter\ifx\csname urlstyle\endcsname\relax
  \providecommand{\doi}[1]{doi: #1}\else
  \providecommand{\doi}{doi: \begingroup \urlstyle{rm}\Url}\fi

\bibitem[Andoni \& Indyk(2008)Andoni and Indyk]{andoni2008near}
Alexandr Andoni and Piotr Indyk.
\newblock Near-optimal hashing algorithms for approximate nearest neighbor in
  high dimensions.
\newblock \emph{Communications of the ACM}, 51\penalty0 (1):\penalty0 117--122,
  2008.

\bibitem[Ardin et~al.(2016)Ardin, Peng, Mangan, Lagogiannis, and
  Webb]{ardin2016using}
Paul Ardin, Fei Peng, Michael Mangan, Konstantinos Lagogiannis, and Barbara
  Webb.
\newblock Using an insect mushroom body circuit to encode route memory in
  complex natural environments.
\newblock \emph{PLoS computational biology}, 12\penalty0 (2):\penalty0
  e1004683, 2016.

\bibitem[Dasgupta \& Tosh(2020)Dasgupta and Tosh]{dasgupta2020expressivity}
Sanjoy Dasgupta and Christopher Tosh.
\newblock Expressivity of expand-and-sparsify representations.
\newblock \emph{arXiv preprint arXiv:2006.03741}, 2020.

\bibitem[Dasgupta et~al.(2017)Dasgupta, Stevens, and
  Navlakha]{dasgupta2017neural}
Sanjoy Dasgupta, Charles~F Stevens, and Saket Navlakha.
\newblock A neural algorithm for a fundamental computing problem.
\newblock \emph{Science}, 358\penalty0 (6364):\penalty0 793--796, 2017.

\bibitem[Indyk \& Motwani(1998)Indyk and Motwani]{indyk1998approximate}
Piotr Indyk and Rajeev Motwani.
\newblock Approximate nearest neighbors: towards removing the curse of
  dimensionality.
\newblock In \emph{Proceedings of the thirtieth annual ACM symposium on Theory
  of computing}, pp.\  604--613, 1998.

\bibitem[Lin et~al.(2014)Lin, Bygrave, De~Calignon, Lee, and
  Miesenb{\"o}ck]{lin2014sparse}
Andrew~C Lin, Alexei~M Bygrave, Alix De~Calignon, Tzumin Lee, and Gero
  Miesenb{\"o}ck.
\newblock Sparse, decorrelated odor coding in the mushroom body enhances
  learned odor discrimination.
\newblock \emph{Nature neuroscience}, 17\penalty0 (4):\penalty0 559--568, 2014.

\bibitem[Shen et~al.(2021)Shen, Xia, Li, Martín-Martín, Fan, Wang,
  Pérez-D’Arpino, Buch, Srivastava, Tchapmi, Tchapmi, Vainio, Wong, Fei-Fei,
  and Savarese]{shenigibson}
Bokui Shen, Fei Xia, Chengshu Li, Roberto Martín-Martín, Linxi Fan, Guanzhi
  Wang, Claudia Pérez-D’Arpino, Shyamal Buch, Sanjana Srivastava, Lyne
  Tchapmi, Micael Tchapmi, Kent Vainio, Josiah Wong, Li~Fei-Fei, and Silvio
  Savarese.
\newblock igibson 1.0: A simulation environment for interactive tasks in large
  realistic scenes.
\newblock In \emph{2021 IEEE/RSJ International Conference on Intelligent Robots
  and Systems (IROS)}, pp.\  7520--7527, 2021.
\newblock \doi{10.1109/IROS51168.2021.9636667}.

\bibitem[Stone et~al.(2017)Stone, Webb, Adden, Weddig, Honkanen, Templin,
  Wcislo, Scimeca, Warrant, and Heinze]{stone2017anatomically}
Thomas Stone, Barbara Webb, Andrea Adden, Nicolai~Ben Weddig, Anna Honkanen,
  Rachel Templin, William Wcislo, Luca Scimeca, Eric Warrant, and Stanley
  Heinze.
\newblock An anatomically constrained model for path integration in the bee
  brain.
\newblock \emph{Current Biology}, 27\penalty0 (20):\penalty0 3069--3085, 2017.

\bibitem[Wystrach et~al.(2020)Wystrach, Le~Mo{\"e}l, Clement, and
  Schwarz]{wystrach2020lateralised}
Antoine Wystrach, Florent Le~Mo{\"e}l, Leo Clement, and Sebastian Schwarz.
\newblock A lateralised design for the interaction of visual memories and
  heading representations in navigating ants.
\newblock \emph{bioRxiv}, 2020.

\end{thebibliography}
\bibliographystyle{iclr2021_conference}

\end{document}